\documentclass[preprint,12pt]{elsarticle}

\usepackage{amssymb}
\usepackage{amsmath}
\input{abrege}
\newsavebox{\fminibox}
\newlength{\fminilength}
\newenvironment{fminipage}[1][\linewidth]
  {\setlength{\fminilength}{#1}%-2\fboxsep-2\fboxrule}%
   \begin{lrbox}{\fminibox}\begin{minipage}{\fminilength}}
  {\end{minipage}\end{lrbox}\noindent\fbox{\usebox{\fminibox}}}

  \def\+{^\dagger}
\def\nequiv{\not\kern-.05em\equiv}
\def\egal{\kern-.5em=\kern-.5em}        % Moins d'espace autour de "="
\def\propt{\kern-.2em\propto\kern-.2em} % Idem
\def\intdouble{\int\kern-0.3em\int}
\def\inttriple{\int\kern-0.3em\int\kern-0.3em\int}

\def\rond#1{\overset{\kern-0.33em~_\circ}{#1}}
\def\rondit[#1]#2{\overset{\kern#1~_\circ}{#2}}

\def\babs{\begin{abstract}}             \def\eabs{\end{abstract}}
\def\barr{\begin{array}}                \def\earr{\end{array}}
\def\bcc{\begin{center}}                \def\ecc{\end{center}}

\def\bdes{\begin{description}}          \def\edes{\end{description}}
\def\bdoc{\begin{document}}             \def\edoc{\end{document}}
\def\ben{\begin{enumerate}}             \def\een{\end{enumerate}}
\def\beqn{\begin{eqnarray}}             \def\eeqn{\end{eqnarray}}
\def\beqnl#1{\beqn\label{#1}}           \def\eeqnl#1{\label{#1}\eeqn}
\def\beqnx{\begin{eqnarray*}}           \def\eeqnx{\end{eqnarray*}}
\def\bseqn{\begin{subeqnarray}}         \def\eseqn{\end{subeqnarray}}
\def\beq#1\eeq{\begin{equation}#1\end{equation}}
\def\bal#1\eal{\begin{align}#1\end{align}}
\def\balx#1\ealx{\begin{align*}#1\end{align*}}
\def\beqx{$$}                           \def\eeqx{$$}
\def\bfig{\protect\begin{figure}}       \def\efig{\protect\end{figure}}
\def\bfigx{\protect\begin{figure*}}     \def\efigx{\protect\end{figure*}}
\def\bfigt{\protect\begin{figurette}}   \def\efigt{\protect\end{figurette}}
\def\bfl{\begin{flushleft}}             \def\efl{\end{flushleft}}
\def\bfr{\begin{flushright}}            \def\efr{\end{flushright}}
\def\bit{\begin{itemize}}               \def\eit{\end{itemize}}
\def\bmi{\begin{minipage}}              \def\emi{\end{minipage}}
\def\bfmi{\begin{fminipage}}            \def\efmi{\end{fminipage}}
\def\bpic{\begin{picture}}              \def\epic{\end{picture}}
\def\bqu{\begin{quote}}                 \def\equ{\end{quote}}
\def\bqun{\begin{quotation}}            \def\equn{\end{quotation}}
\def\bsl{\begin{slide}}                 \def\esl{\end{slide}}
\def\btabb{\begin{tabbing}}             \def\etabb{\end{tabbing}}
\def\btabl{\begin{table}}               \def\etabl{\end{table}}
\def\btablx{\begin{table*}}             \def\etablx{\end{table*}}
\def\btab{\begin{tabular}} %\def\etab{\end{tabular}} Conflit avec eta gras... Elle est pas grasse, ma soeur !
\def\btabu{\begin{tabular}}             \def\etabu{\end{tabular}}
\def\btabx{\begin{tabular*}}            \def\etabx{\end{tabular*}}
\def\bbib{\begin{thebibliography}{999}} \def\ebib{\end{thebibliography}}
\def\bver{\begin{verbatim}}             \def\ever{\end{verbatim}}
\def\bca{\begin{cases}}                          \def\eca{\end{cases}}

\def\addNN#1{{\noindent \color{red}{#1}}}

\usepackage{color}

\def\rev#1{{\color{red}{#1}}}
\def\revfinal#1{{\color{blue}{#1}}}

\input{alphabet}

\usepackage{subcaption}
\usepackage{hyperref}
\usepackage{booktabs}
\usepackage{adjustbox}
\usepackage{multirow}
\usepackage{hhline}
\usepackage{pifont}
\newcommand{\ceckmark}{\ding{51}}%
\newcommand{\xmark}{\ding{55}}%

\newcommand{\smallsim}{\smallsym{\mathrel}{\sim}}

\makeatletter
\newcommand{\smallsym}[2]{#1{\mathpalette\make@small@sym{#2}}}
\newcommand{\make@small@sym}[2]{%
  \vcenter{\hbox{$\m@th\downgrade@style#1#2$}}%
}
\newcommand{\downgrade@style}[1]{%
  \ifx#1\displaystyle\scriptstyle\else
    \ifx#1\textstyle\scriptstyle\else
      \scriptscriptstyle
  \fi\fi
}
\makeatother

\begin{document}

\begin{frontmatter}


% \author{Name\corref{cor1}\fnref{label2}}
% \affiliation{organization=}

%\title{}
\title{Cross-Attention Fusion of Visual and Geometric Features for Large Vocabulary Arabic Lipreading}

\author[label1]{Samar Daou\corref{cor1}}\ead{samar.daou@enis.tn}
\author[label1,label2]{Achraf Ben-Hamadou}\ead{achraf.benhamadou@crns.rnrt.tn}
\author[label1,label3]{Ahmed Rekik}\ead{ahmed.rekik@crns.rnrt.tn}
\author[label1,label2]{Abdelaziz Kallel}\ead{abdelaziz.kallel@crns.rnrt.tn}

\affiliation[label1]{organization={SMARTS Laboratory},addressline= {Technopark of Sfax},  city={Sakiet Ezzit}, postcode={3021}, city={Sfax}, country={Tunisia}}
%\affiliation[label1]{organization={Laboratory of Signals, systeMs, aRtificial Intelligence and neTworkS},addressline= {Technopark of Sfax},  city={Sakiet Ezzit}, postcode={3021}, city={Sfax}, country={Tunisia}}
% %%
\affiliation[label2]{organization={Digital Research Center of Sfax},addressline= {Technopark of Sfax},  city={Sakiet Ezzit}, postcode={3021}, city={Sfax}, country={Tunisia}}

\affiliation[label3]{organization={ISSAT, Gafsa university},addressline= {Sidi Ahmed Zarrouk University Campus},  city={Gafsa}, postcode={2112}, country={Tunisia}}

\cortext[cor1]{Corresponding author}
% \author{} %% Author name

% \affiliation{organization={},%Department and Organization

%% Abstract
\begin{abstract}
Lipreading involves using visual data to recognize spoken words by analyzing the movements of the lips and surrounding area. It is a hot research topic with many potential applications, such as human-machine interaction and enhancing audio speech recognition.
Recent deep-learning based works aim to integrate visual features extracted from the mouth region with landmark points on the lip contours. However, employing a simple combination method such as concatenation may not be the most effective approach to get the optimal feature vector. To address this challenge, firstly, we propose a cross-attention fusion-based approach for large lexicon Arabic vocabulary to predict spoken words in videos. Our method leverages the power of cross-attention networks to efficiently integrate visual and geometric features computed on the mouth region.
Secondly, we introduce the first large-scale Lipreading in the Wild for Arabic (LRW-AR) dataset containing 20,000 videos for 100-word classes, uttered by 36 speakers.
The experimental results obtained on LRW-AR and ArabicVisual databases showed the effectiveness and robustness of the proposed approach in recognizing Arabic words. Our work provides insights into the feasibility and effectiveness of applying lipreading techniques to the Arabic language, opening doors for further research in this field. Link to the project page: \href{https://crns-smartvision.github.io/lrwar}{https://crns-smartvision.github.io/lrwar}
\end{abstract}

% \includegraphics[width=1\textwidth]{abstract2.pdf}

% \begin{highlights}

% \item A novel cross-attention fusion technique for integrating visual and geometric features in large vocabulary Arabic lipreading system.
% \item First large-scale lipreading dataset in Arabic, designed for word-level prediction.
% \item Achieved high accuracy using a combination of visual and geometric data to enhance lipreading performance.

% \end{highlights}

%% Keywords

\end{frontmatter}




\section{Introduction}\label{sec1}

%context and motivation
Lipreading is the technique for recognizing spoken words by analyzing the movements of the lips and surrounding area of a speaking person in a video. Its applications span across different domains such as assisting deaf people \cite{ivanko2019automatic}, analysis of criminal conversations \cite{rothkrantz2017lip}, speaker identification \cite{lu2019lip} and human-machine interaction systems such as dictating instructions \cite{chowdhury2022lip,sheng2022deep,pat-achraf}. This technique proves especially valuable in challenging environments with noise or multiple speakers, where hearing is challenging, yet comprehension is essential \cite{dhanjal2023comprehensive}. Additionally, when combined with audio, lipreading can offer supplementary information that enhances the speech recognition system's performances, especially in noisy conditions \cite{li2023improving}.

% what makes lipreading challenging
Learning to read lips is challenging due to various factors. One main challenge is the similarity in lip movements for certain distinct characters in a language, such as 'p' and 'b' in English. Furthermore, lipreading encounters additional challenges, such as variations in facial appearance, accents, speaking speed and manner, as well as variations in facial pose \cite{akhter2022diverse}. These factors further contribute to the complexity of lipreading and require robust training and advanced architectures to account for individual differences and enhance accuracy in lipreading systems. 

%Related Work limitations
In recent years, there has been growing interest in the research field of lipreading, mainly driven by the advancements in deep learning techniques and the availability of parallel computation acceleration hardware. Deep learning-based lipreading systems have demonstrated superior performance compared to earlier approaches based on handcrafted image descriptors for many applications \cite{fenghour2021deep}.

Deep learning-based lipreading systems encounter several challenges. It is a fact that the availability of learning datasets significantly influences the development of deep learning network architectures and language-specific lipreading systems, considering the substantial variations in pronunciation, phonemes, and lip movements between languages \cite{jitaru2021toward}. In this sense, several lipreading datasets have been created in different languages to fulfill this requirement. For instance, an English lipreading dataset \cite{lai_lip_2017} comprises 500,000 instances of spoken words, while a Mandarin dataset \cite{yang_lrw-1000_2019} contains 286 Chinese syllables represented by more than 1 million character occurrences. More recently, a Russian dataset \cite{egorov_lrwr_2021} was introduced, consisting of over 117,500 samples of 235 words. In this work, we specifically focus on the Arabic language. Despite efforts made in this field, it is worth noting that Arabic, spoken by over 274 million people worldwide \footnote{https://www.statista.com/statistics/266808/the-most-spoken-languages-worldwide/, visited on September 10th, 2024}, lacks an existing large-scale lipreading dataset or a dedicated system designed to this language.

Moreover, using only visual features computed on the cropped mouth region for lipreading has big challenges. 
These features include a lack of contextual information, inherent ambiguity and variability in lip patterns, limited discriminative power, and sensitivity to environmental factors. However, these limitations can be effectively mitigated by incorporating facial landmarks associated with video frames. By explicitly incorporating facial landmarks, lipreading systems gain access to additional precise geometric information regarding fine-grained lip movements and shape variations, which enables better differentiation of visually similar phonemes. While some existing methods use both video frames and landmarks for lipreading (\eg \cite{sheng2021adaptive,visapp23}), they often rely on basic combination methods such as feature concatenation that may limit their ability to effectively capture essential mutual information. In this work, we focus specifically on the fusion aspect and introduce a sophisticated cross-attention mechanism for optimally combining visual and landmark features in our lipreading system.

%Contribution: What idea do you have to address Related Work limitations?
To summarize, our main contributions are as follows:
\begin{itemize}
    \item We introduce an efficient cross-attention fusion mechanism that merges image and facial landmarks for lipreading. Our fusion strategy aims to capture a larger visual context and improve the combination of mouth appearance and landmark features.
    
    \item We present the first large-scale Arabic dataset for word-level lipreading containing 20,000 videos for 100 different word classes.
\end{itemize} 

% paper organization
The remainder of this paper is organized as follows: First, we provide an overview of related works. Afterward, we outline the architecture of our lipreading system, followed by a presentation of the proposed dataset, including details on data collection, annotation, and preprocessing methods. We then report the obtained experimental results. Finally, we conclude this paper by summarizing our findings and highlighting areas for future work.

%----------------------------------------------------------------------------
\section{Related work}\label{sec2}
In this section, we start with an overview of publicly available lipreading datasets in multiple languages, with a specific focus on the Arabic language. Then, we review the most relevant deep learning-based lipreading approaches closely related to our work.

\subsection{Lipreading datasets}
\paragraph{Large-scale lipreading datasets}
Despite many efforts to improve classification performance, lipreading networks are known for requiring extensive training data. Consequently, early attempts at deep learning-based methods faced a data shortage, leading to the increased popularity of databases with more samples per class. Table \ref{table:ps} summarizes the most popular publicly available datasets.

The GRID dataset, introduced in 2006 \cite{cookea_audio-visual_2006}, was a pioneering attempt to construct a large-scale English lipreading dataset in terms of the number of samples per class but had limited vocabulary. 
To address this limitation, new databases have been created more recently to provide a more extensive vocabulary and a larger number of samples.
\begin{table}[h]
\centering
\small

\centering
\begin{adjustbox}{max width=1\textwidth}
\begin{tabular}{@{}llllllll@{}}
\toprule
Dataset & Language & Speakers & Task & Classes & Utterances & Duration & Year\\
\midrule
GRID \cite{cookea_audio-visual_2006} & English & 34 & Sentences & 51 & 34,000 & $\smallsim$ 27 h & 2006\\
LRW \cite{lai_lip_2017} & English & 1,000+ & Words & 500 & 400,000 & $\smallsim$111 h & 2016\\
VoxCeleb \cite{nagrani_voxceleb_2017} & English & 2000+ & Sentences & 1000 & - & - & 2017\\
LRS \cite{chung2017lip} & English & 1,000+ & Sentences & 17,428 & 118,116 & $\smallsim$ 33 h & 2017\\
MV-LRS \cite{profile} & English & 1,000+ & Sentences & 14,960 & 74,564 & $\smallsim$ 20 h & 2017\\
LRS2-BBC \cite{afouras_deep_2018} & English & 1,000+ & Sentences & - & 8 M & - & 2017\\
LRS3-TED \cite{afouras_lrs3-ted_2008} & English & 5000+ & Sentences & - & 1,2 M & $\smallsim$ 400 h & 2018\\
VoxCeleb2 \cite{chung_voxceleb2_nodate} & English & 6000 & Sentences & - & 1 M & - & 2018 \\
LRW-1000 \cite{yang_lrw-1000_2019} & Mandarin & 2000+ & Words & 1000 & - & - & 2018\\
VoxSRC \cite{chung_voxsrc_nodate} & English & 1000+ & Sentences & - & 19,154 & - & 2019 \\
LRWR \cite{egorov_lrwr_2021} & Russian & 135 & Words & 235 & - & - & 2021\\
\bottomrule
\end{tabular}
\end{adjustbox}
\caption {Most known large vocabulary lipreading datasets.}

\label{table:ps}
\end{table}

The most relevant among them are the LRW \cite{lai_lip_2017}, LRS \cite{chung2017lip}, and MV-LRS \cite{profile} databases. The LRW and LRS databases are derived from BBC program recordings between 2010 and 2016.

LRW dataset \cite{lai_lip_2017} contains sentences from more than 1,000 speakers and a vocabulary of 500 words that appear at least 800 times each (400,000 in total). Therefore, the large number of speakers helps to generalize unseen speakers. LRS dataset \cite{chung2017lip} consists of 17,428 distinct words combined in 118,116 statements, as well as the corresponding facial track. Also, MV-LRS database \cite{profile} has been recorded from BBC programs; in contrast to LRW and LRS which only include front views, MV-LRS contains images from multiple viewing angles ranging from 0 to 90 degrees. In addition, LRS2-BBC \cite{afouras_deep_2018} dataset is like an extension to LRS that consists of more than 8 million utterances. LRS3-TED \cite{afouras_lrs3-ted_2008} dataset includes more than 400 hours of video, taken from 5594 TED and TEDx talks in English. 

LRW-1000 \cite{yang_lrw-1000_2019} is a naturally distributed lipreading dataset that contains 1,000 classes with approximately 1.14 million Chinese character instances and spans 286 Chinese syllables. This database is the largest and the only publicly available Mandarin lipreading dataset at the moment. LRWR dataset \cite{egorov_lrwr_2021} is the largest available dataset in the Russian language for visual speech recognition and contains 235 classes and 135 speakers. While, VoxCeleb \cite{nagrani_voxceleb_2017}, VoxCeleb2 \cite{chung_voxceleb2_nodate}, and VoxSRC \cite{chung_voxsrc_nodate} are publicly available datasets, they consist of short clips of videos taken from unconstrained YouTube videos of thousands of celebrities talking in the wild.

\paragraph{Lipreading datasets for Arabic language}

%Arabic
Compared to other languages, the unavailability of large-scale datasets for Arabic has been a barrier to developing efficient lipreading systems for the Arabic language. Table \ref{table:arabic_table} lists the few attempts to construct lipreading datasets for Arabic. 

The AVAS dataset \cite{Sagheer2013AudioVisualAS} was introduced in 2013 as the first Arabic dataset in the field. It includes recordings of 50 speakers uttering digits, words, and phrases. 
\begin{table*}[h]
\centering
\begin{adjustbox}{max width=1\textwidth}
\begin{tabular}{@{}lllllll@{}}
\toprule
Name & Year & Speakers & Task & Classes & Utterances & Availability\\
\midrule
AVAS \cite{Sagheer2013AudioVisualAS} & 2013  & 50 & Digits, words, phrases & 48 & - & \xmark \\
RML \cite{dweik2022read} & 2022 & 73 & Words & 10 & - & \xmark \\
ArabicVisual \cite{alsulami2022deep} & 2022  & 22 & Digits, phrases & 14 & 2400 & \checkmark \\
\bottomrule
\end{tabular}
\end{adjustbox}
\caption{List of available lipreading datasets for Arabic.}
\label{table:arabic_table}
\end{table*}
However, this dataset lacks information on the number of classes or utterances included, limiting its utility for researchers. In 2022, the RML dataset \cite{dweik2022read} was recorded with 73 speakers and included 10 classes. 
However, the exact number of utterances in the RML dataset is not specified. More recently, the ArabicVisual dataset \cite{alsulami2022deep} was released, involving 24 speakers and 14 words, for a total of 2400 utterances. While this dataset represents a significant improvement, similar to the AVAS and ArabicVisual datasets, the RML dataset is still limited in terms of the number of words and speakers.

\subsection{Lipreading methods}\label{subsec:LR_Methods}
The number of works addressing the lipreading task has increased significantly over the last decade, thanks to large-scale datasets availability and developments in deep learning techniques \cite{malhotra2023recent}. In the following sections, we begin by highlighting key works dedicated to the Arabic language, followed by an overview of lipreading methods designed for other languages.

\paragraph{Lipreading methods for Arabic language}
Recent works on lipreading for the Arabic language have mainly focused on word recognition. One such system is the system proposed by Dweik \etal \cite{dweik2022read}, which is a lipreading system employing deep learning techniques such as convolutional neural networks (CNN), time-delayed CNN with long short-term memory (TD-CNN-LSTM), and time-delayed CNN with bidirectional LSTM (TD-CNN-BiLSTM). When tested on the RML dataset, they achieved an overall prediction accuracy of 82.84\% on test data using the CNN model with RGB dataset.

Additionally, Alsulami \etal \cite{alsulami2022deep} proposed their Arabic lipreading approach based on VGG-19 as a backbone model to extract visual features. The proposed approach achieved an accuracy of 94\% for digit recognition, 97\% for sentence recognition, and 93\% for digits and sentences recognition in the ArabicVisual dataset.

\paragraph{Lipreading methods for other languages}
Several studies have explored different approaches for visual speech recognition, mainly focusing on the English language.

Gutierrez \etal \cite{gutierrez2017lip} presented some word prediction models based on the MIRACL-VC1 dataset \cite{rekik2014new}. The data was processed by detecting and extracting the subject's face area in each video frame before entering the entire video sequence into the model. Their proposed architecture is based on deep multi-layer CNN models and LSTM networks, taking inspiration from LipNet \cite{assael_lipnet_2016}. The extensive experimental results demonstrated the importance of dropout, data augmentation, hyperparameter tuning, including visible and invisible validation splits, and batch normalization for the regularization of these models.

The lipreading approach proposed by Stafylakis et al. \cite{stafylakis_combining_nodate} consists of a 3D convolutional neural network and a residual network designed to capture relevant visual data from the input sequence for each timestamp. These representations are then passed into a bidirectional LSTM. The network was trained and tested on the LRW dataset using various configurations, with the best configuration achieving an accuracy of 83.0\%, outperforming previous works presented in \cite{lai_lip_2017,profile}.

The multi-tower structure is another significant advancement in the field introduced in \cite{chung2018learning}. This architecture involves multiple towers, with each tower accepting either a single frame or a T-channel image, where each channel corresponds to an individual grayscale frame. The activation outputs from all these towers are then merged to yield the ultimate representation of the entire sequence. This multi-tower approach has demonstrated remarkable success, particularly on the LRW dataset, with 95.6\% accuracy.

Ma \etal \cite{ma_towards_2021} have proposed a lipreading model that is specifically designed for recognizing isolated words. Their model is composed of a 3D convolutional network, 18 layers of residual network and a temporal convolutional network (TCN). The proposed model has achieved prominent results on the LRW and LRW-1000 datasets. Recently, the same authors introduced a new depth temporal convolutional layer head in \cite{martinez_lipreading_2020}, which improves the performance of the model with a significant reduction in computational cost. With this updated architecture, they achieved state-of-the-art results with an accuracy of 88.6\% and 46.6\% on the LRW and LRW-1000 datasets, respectively.

The existing deep learning techniques for lipreading primarily focus on exploiting appearance and optical flow features from videos. However, these methods do not fully utilize the potential of lip motion characteristics. In addition to appearance and optical flow, the dynamic mouth contour conveys valuable complementary information. Unfortunately, the modeling of dynamic mouth contours has not received sufficient attention compared to appearance and optical flow.

To bridge this gap, \cite{sheng2021adaptive} proposed an Adaptive Semantic-Spatio-Temporal Graph Convolution Network (ASST-GCN). By leveraging the architecture of graph convolutional networks, ASST-GCN captures the inherent relationships between the points of the mouth contour and their temporal dynamics. To combine the complementary information from appearance and mouth contour, they introduce a two-stream visual front-end network that simultaneously processes both types of data to extract distinctive features. And lately, Daou \etal \cite{visapp23} proposed a two-stream deep learning architecture. This architecture aims to integrate visual and geometric features extracted from the lips and surrounding area. However, while the fusion of information from the two previous works \cite{sheng2021adaptive,visapp23} is straightforward and relies on basic concatenation, it may limit its efficiency in capturing mutual information between the merged streams. 

\paragraph{Attention mechanisms for lipreading}
Recently, attention mechanisms have gained significant attraction in different domains \cite{naderi2023cross} due to their ability to focus on informative parts of the input data. In the context of lipreading self-attention mechanism has been explored for modeling temporal relationships within an input video sequence. For instance, Lohrenz \etal \cite{lohrenz2023relaxed} propose a self-attention based architecture for lipreading, enabling the model to focus on  relevant features extracted across the video sequence.
Differently, cross-attention mechanisms have shown promising results in different applications and domains, especially for leveraging complementary additional modalities.

Nagrani \etal \cite{nagrani2018seeing} applied cross-attention mechanisms in their multi-modal speech recognition system. Their approach showed significant improvements in recognizing speech under challenging conditions by effectively combining audio and visual signals, highlighting the power of cross-attention in enhancing speech recognition.

In natural language processing, Tsai \etal \cite{tsai2019multimodal} proposed a multi-modal transformer network that employs cross-attention to fuse audio, visual, and textual information. Their model achieved state-of-the-art results on several multi-modal benchmarks, showcasing the effectiveness of cross-attention in handling diverse data sources.

Furthermore, in their work Rajan \etal \cite{rajan2022cross} investigated the use of cross-attention for multi-modal emotion recognition. They found that cross-attention mechanisms can outperform self-attention by better capturing the interactions between visual and auditory features. This results in an improvement of the accuracy of emotion recognition systems, underscoring the potential of cross-attention in effectively merging different types of data.

\subsubsection{Our method positioning} 

To define lip movement, we propose a model that takes advantage of cross-modal properties.
It employs a cross-attention mechanism to effectively align visual data with facial landmarks. This alignment enhances the fusion of diverse features, resulting in improving the model's performance. 
This approach contrasts with single attention fusion, which may not fully capture the complex dependencies and interactions between modalities. Although previous works using single attention mechanisms, such as Naderi \etal \cite{naderi2023cross}, have shown potential, our results reveal that the cross-attention offers a significant performance enhancement.

To deliver high performance, we propose employing convolutional neural networks, including 3D and 2D convolutional models for extracting visual features and Graph Neural Networks to extract geometric features, along with Temporal Convolutional Networks that has been highly effective in recent deep learning-based lipreading systems. 

Motivated by this success in many languages, we propose the integration of these techniques into Arabic lipreading systems and study its potential. This adaptation could considerably improve the performance of computer-assisted communication applications in this language.

%----------------------------------------------------------------------------
\section{Proposed approach}\label{sec3}
\begin{figure*}[t]
    \centering
    \includegraphics[width=1\textwidth]{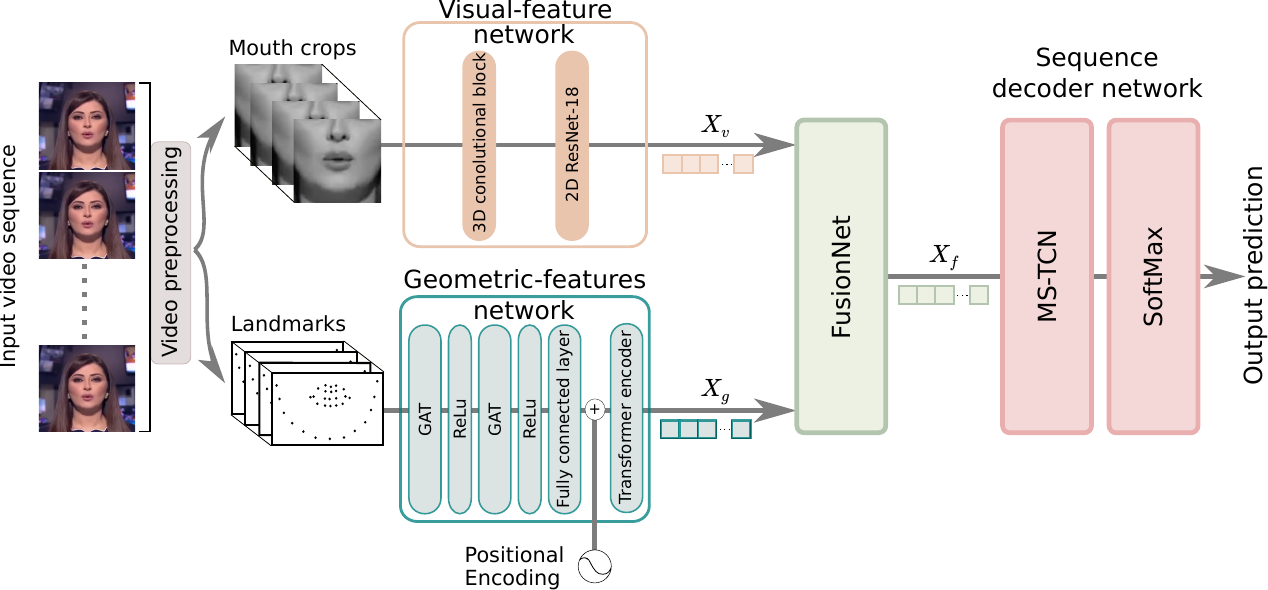}
    
  \caption{Outline of the proposed approach. Video preprocessing: crop the mouth region from the input video sequence and obtain the corresponding facial landmarks. Visual-feature network: extract relevant information from the preprocessed data. Geometric-feature network: encodes lip contour variation delivered by facial landmarks. FusionNet network: fuses the encoded features. Sequence back-end network: based on a multi-scale temporal convolutional network (MS-TCN) to encode temporal variation and classify the input video sequence.}
    \label{fig:flowchart}
\end{figure*}

As depicted in Fig. \ref{fig:flowchart}, our approach consists of five key steps. The first step involves video preprocessing, wherein we crop the mouth region of $T$ frames from the input video sequence and acquire corresponding facial landmarks. This step yields both visual and geometric data streams. Afterward, a Visual-feature network is employed to extract a sequence of $T$ feature vectors for the visual data stream,  and a geometric-feature network to encode the variation in lips contour delivered by facial landmarks. Moving forward, a FusionNet network is used to combine these two sequences into a single sequence of $T$ feature vectors. The final step outputs the predicted speech by applying a sequence decoder network. Detailed descriptions of each module involved in these steps are provided below.

\subsection{Input data preprocessing}
Assuming that a face detector is first applied to the input video to broadly localize the talking face in all input frames, we applied a generic facial landmark detector \cite{sagonas2013300} to identify standard 68 landmarks on every input frame. Other landmark detection methods could be also considered for this study, as long as they cover landmarks on the lips. We use these landmarks to align and resize the frames, resulting in a mouth region of $96 \times 96$ pixels to ensure its position is nearly centered in the extracted images. Additionally, these cropped patches are converted to grayscale and normalized by subtracting the mean and dividing by the standard deviation values. As a data augmentation process during the training phase, an additional cropping is applied to obtain 88$\times$88 size patches at random positions. For the validation and testing phases, a central cropping of 88$\times$88 size patches is applied instead. On the other hand, the extracted landmarks undergo the same affine transformations used for frame preprocessing to ensure they are represented in the same coordinate system as the extracted image patches, thereby geometrically aligning the patches and landmarks. This preprocessing step yields two streams of data: the extracted images of the mouth region, and the sequence of extracted facial landmarks.

\subsection{Visual-feature network}
The visual feature network encodes the appearance and movement of the lips over the extracted image sequence. It consists of a single 3D convolutional block followed by a modified 2D ResNet-18 architecture \cite{he2016deep}. The 3D convolutional layer takes as input a sequence of $T$ consecutive grayscale images of size $96 \times 96$. It is a sequence of 64 kernels of size $5\times7\times7$, followed by batch normalization, ReLU activation, and max pooling with a kernel size of $1 \times 3 \times 3$. This initial network outputs spatio-temporal features with dimensions of $T \times 64 \times 24 \times 24$. 
These spatio-temporal features are then fed into the ResNet-18 architecture, which consists of four stages of convolutional blocks. Each stage contains several residual blocks, with the number of feature channels progressively increasing from 64 in the first stage, to 128 in the second, 256 in the third, and finally 512 in the fourth stage. This hierarchical feature extraction process enables the network to capture increasingly complex patterns and representations of the lip movements. After passing through the ResNet-18 blocks, the features undergo adaptive average pooling, which reduces each feature map to a single value, resulting in a feature vector. The final output of the visual feature network is the visual features $\Xb_v \in \mathbb{R}^{T\times d}$, where $d=512$. These features serve as a compact and informative representation of the visual input, capturing both spatial and temporal dynamics crucial for the subsequent stages of lipreading.

\subsection{Geometric-feature network}
Similar to the visual feature network, the geometric feature network encodes the appearance and movement of the lips. However, the geometric features are extracted based on an explicit representation of the mouth region using the sequence of extracted landmarks. Let us define $N$ as the number of used landmarks per frame in this network. The input size of the geometric feature network is $T\times N\times 2$, as each landmark is represented by its 2D coordinates. 

To encode the spatial arrangement between the facial landmarks, each frame represents a graph, and each landmark is a node within the graph. We applied the K-nearest neighbors algorithm to create the edges between the nodes, with $k$ set to $5$. The network includes two Graph Attention Network (GAT) \cite{velivckovic2017graph} layers: the first layer transforms input graphs from 2 to 16 channels per node, and the second layer further expands it to 64 channels, followed by a fully connected layer to output a feature vector of $512$ per graph. Moreover, to encode the temporal variations of landmarks over the frames, we consider a positional encoding of size $512$ to be added to the $T$ vectors and fed into the sequence $T \times 512$, which is then processed by a transformer encoder block of a single layer with 8 heads and $1024$ as a dimension of feed-forward network model, leading to the geometric feature tensor $\Xb_g \in \mathbb{R}^{T\times d}$ of the same size as the visual features.

\begin{figure*}[t]
    \centering
    \includegraphics[width=0.7\textwidth]{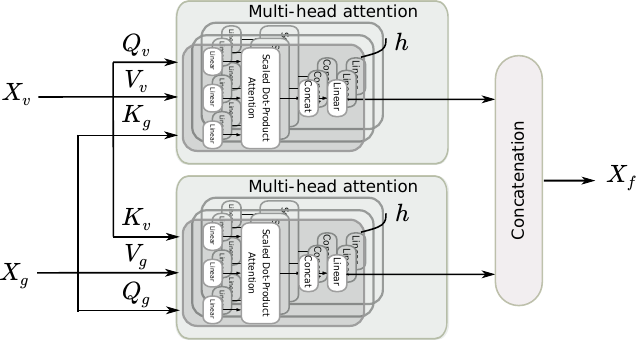}
    \caption{Architecture of the FusionNet: multi-modality cross-attention fusion network.}
    \label{fig:FUSION}
\end{figure*}

\subsection{FusionNet network}
The proposed FusionNet is based on a multi-head attention mechanism to aggregate the fused features from both the visual and geometric features extracted from their corresponding networks. We keep the original notation as in the original description of the Transformer architecture \cite{vaswani2017attention}. As illustrated in Fig. \ref{fig:FUSION}, it consists of two multi-head attention modules that exchange their $\Kb$ (key) features. The idea behind this design is to integrate features from both modalities. After applying the multi-head attention unit involving $h$ heads, we concatenate the obtained features into one single feature map, noted by $\Xb_f \in \mathbb{R}^{T \times d_{f}}$ with $d_{f}=1024$. The multi-modal fusion process is mathematically represented in Equation \ref{eq:cross_modal_attention}

\begin{equation}
\begin{aligned}
\text{$\Xb_f$} &= \text{Concatenation} \big(\text{Attend}_{\text{V}}(X_{\text{V}}, X_{\text{G}}),
\text{Attend}_{\text{G}}(X_{\text{G}}, X_{\text{V}}) \big)
\end{aligned}
\label{eq:cross_modal_attention}
\end{equation}

\noindent where, $\Xb_f$ represents the enriched feature representation resulting from the fusion process, incorporating two distinct multi-head attention mechanisms dedicated to cross-modal interactions:

\begin{itemize}
    \item \(\text{Attend}_{\text{V}}(X_{\text{V}}, X_{\text{G}})\) represents the attention mechanism applied to the "Vision" input (\(X_{\text{v}}\)) attending to the "Landmark" input (\(X_{\text{G}}\)), capturing interplay of information between modalities.
    \item \(\text{Attend}_{\text{G}}(X_{\text{V}}, X_{\text{G}})\) signifies the attention mechanism applied to the "Landmark" input (\(X_{\text{G}}\)) attending to the "Vision" input (\(X_{\text{V}}\)), facilitating knowledge exchange across modalities.
\end{itemize}

\begin{figure}[t]
    \centering
    \includegraphics[width=1\textwidth]{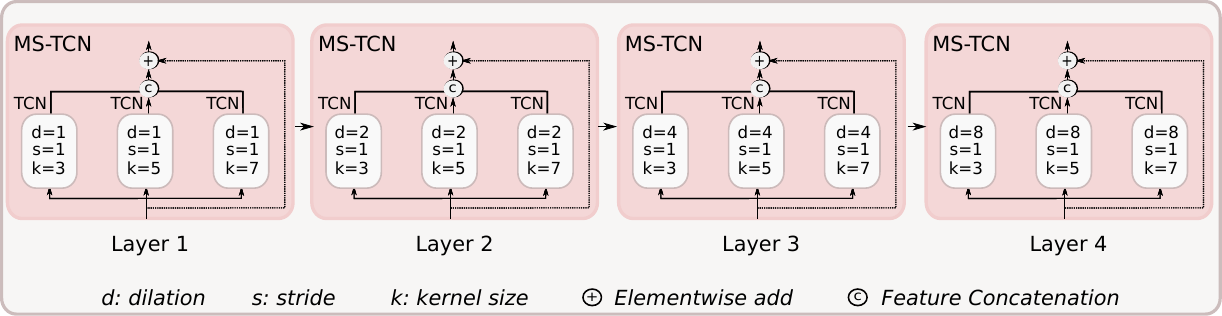}
    \caption{Sequence decoder network with four layers of multi-scale expanded TCN with 1, 2, 4, and 8 dilation values, respectively. Each layer is composed of three  TCN units with 3, 5, and 7 kernel values.}
    \label{fig:mstcn}
\end{figure}
\subsection{Sequence decoder network}

This sequence decoder network is based on the temporal convolution network (TCN) since it has been shown to significantly improve performance on word-level lipreading tasks \cite{ma_towards_2021}. Figure \ref{fig:mstcn} presents a detailed representation of the proposed architecture. It contains multi-scale dilated TCN (MS-TCN) layers, where each layer comprises multiple TCN branches, which are 1D convolutions designed for the temporal domain, with different kernel sizes capturing information across various temporal scales. Our sequence decoder network consists of four layers of MS-TCN with dilation values of 1, 2, 4, and 8, respectively. Each layer is composed of three TCN units using kernel sizes of 3, 5, and 7. The outputs from these branches are concatenated and followed by a fully connected layer and a SoftMax layer, allowing the network to integrate information across multiple temporal scales for an improved spoken word prediction.

%----------------------------------------------------------------------------
\section{LRW-AR dataset}\label{sec4}
\begin{figure}[t]
\centering
\includegraphics[width=0.9\textwidth]{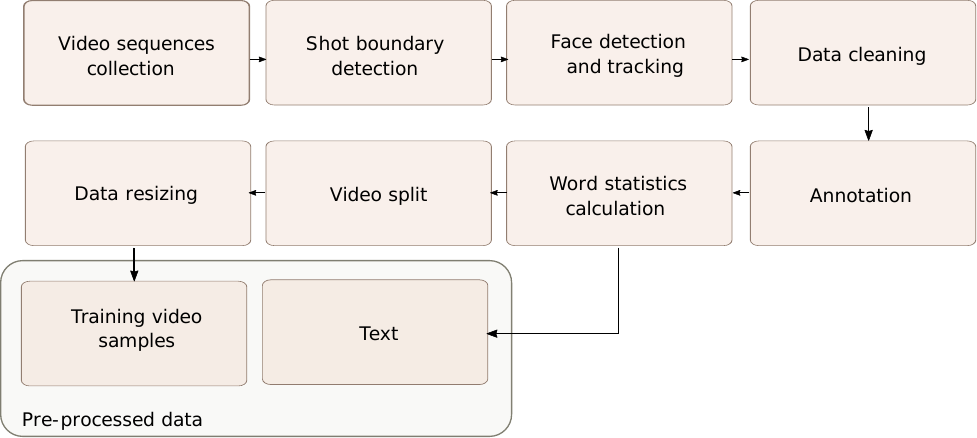}
\caption{Pipeline to generate the LRW-AR dataset.}
\label{fig:pipeline}
\end{figure}

In this section, we present an overview of the procedures followed to collect and prepare our large-scale dataset for Arabic lipreading. The fundamental concept was to gather videos of individuals speaking in Arabic in front of cameras from the YouTube platform, which is a rich source of diverse and natural language data. The process involved several stages, including video scrapping, shot boundary detection, face detection and tracking, data cleaning, annotation, video split, and data cropping, as illustrated in Figure \ref{fig:pipeline}. A diverse set of Arabic videos was collected and pre-processed to extract the frames of interest for lipreading. The frames were then annotated with the corresponding spoken words to create a ground truth for training and evaluation purposes. Finally, the annotations were visually validated to ensure their accuracy and consistency. Each of these steps will be described in detail in the following sections. LRW-AR is released under BY-NC-ND license, and is accessible through the following link: \href{https://osf.io/rz49x}{https://osf.io/rz49x}.

\subsection{Data collection and preprocessing} 
\paragraph{Video scrapping} 
All videos were scraped from YouTube by selecting TV channel playlists that contained Arabic-language TV shows, news programs, and interviews. The playlists were chosen to cover a variety of speakers, genders, and accents.

\paragraph{Shot boundary detection} In this stage, we used a HOG-based scene detection algorithm \cite{king2009dlib} to detect scene changes in the collected videos, and they were split into shorter segments based on scene transitions. This allowed us to focus on the segments where the speaker was present and speaking.

\paragraph{Face detection and tracking:} 
To ensure that our dataset only includes video sequences with a single speaker, the HOG-based face detection method \cite{king2009dlib} was applied to each raw video frame. Then, all face detections belonging to the same person were grouped across frames using a KLT tracker \cite{lucas1981iterative}. This filtering process helps to eliminate any potential confusion that may arise from having multiple speakers in a single video sequence. After filtering, we sample all collected video segments into fixed-length sequences of 1.2 seconds each.

\begin{figure}[t]
\centering\includegraphics[width=1\textwidth]{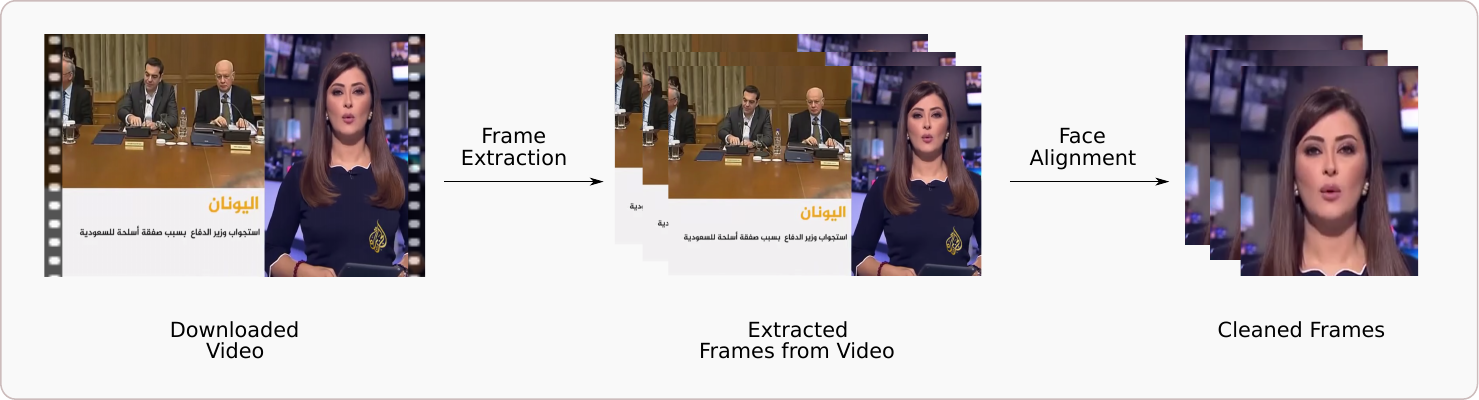}
\caption{Frame extraction and face alignment process.}
\label{fig:frame_extraction}
\end{figure}

\paragraph{Data cleaning:} To ensure the dataset's quality, a manual inspection and cleaning process was carried out, ensuring that only videos featuring a single speaker speaking directly in front of the camera were included.
    
\paragraph{Annotation:} For dataset annotation, we utilized the Vosk speech recognition system\footnote{https://alphacephei.com/vosk/} to automatically transcribe the spoken words in the video sequences into Arabic text. The use of Vosk not only facilitated transcriptions but also established boundaries, indicating the start and end of each spoken word within the text. While we anticipated a perfect alignment between text and audio data, there were some instances where the transcription was incorrect. To rectify this, manual correction of these errors was performed during the final phase of visual validation and dataset quality control.

\paragraph{Video split:} To generate a varied collection of Arabic words, statistical analysis was conducted on the transcriptions to identify the most frequently occurring words. This approach allowed us to focus on the prevalent words in Arabic speech. Subsequently, each video segment was divided into shorter clips, with each clip representing a single word. This procedure facilitated the development of an extensive dataset encompassing a wide array of Arabic words.

\paragraph{Data cropping:} In this step, we standardized all videos to a consistent size of 256x256 pixels. This uniformity enhances overall consistency and simplifies comparisons across different video segments. We cropped each video to focus solely on the speaker's face, minimizing any extraneous information and highlighting the lip movements of the speaker (see Figure \ref{fig:frame_extraction}). This enables us to achieve a more precise and comprehensive understanding of visual characteristics. 

The resulting dataset contains a diverse set of Arabic words spoken by various speakers in different contexts, which can be used for lipreading research and the development of Arabic speech recognition systems. By sharing the methodology, the purpose is to facilitate the creation of similar datasets for other languages and cultures and promote the development of more inclusive visual speech recognition systems.

\begin{figure}[t]
\centering\includegraphics[width=1\textwidth]{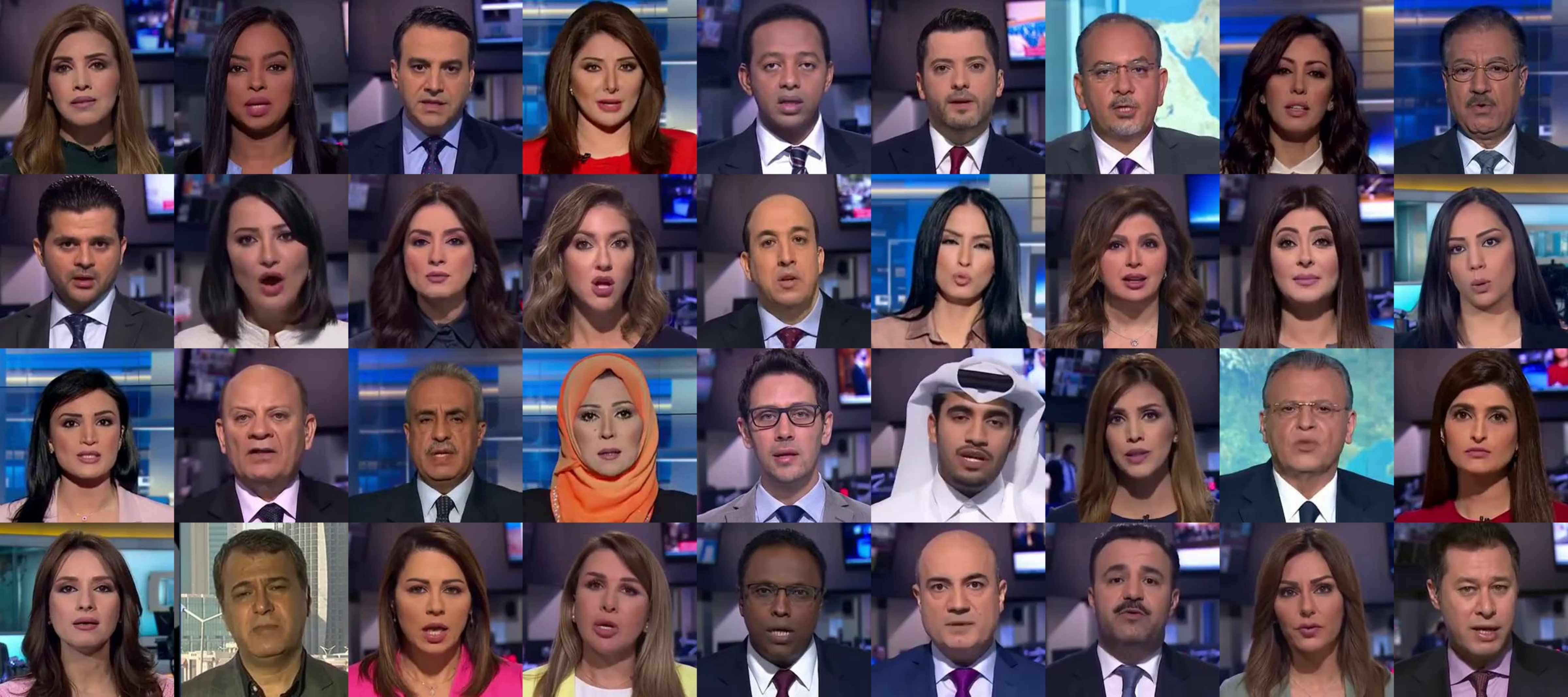}
\caption{A sample of speakers in LRW-AR dataset.}
\label{fig:Jazira_Faces}
\end{figure}

\subsection{Dataset statistics}\label{subsec:data_collection}
The proposed dataset consists of audio-visual data collected from YouTube videos from 2008 until 2022. The dataset contains 100 classes, each associated with a distinct word, and involves a total of 36 speakers (see Figure \ref{fig:Jazira_Faces}). Notably, the speakers are talking spontaneously, presenting additional challenges for lipreading systems, including differences in pronunciation, intonation, variations in face pose, and background noise. Each word was uttered 200 times, resulting in a total of 20,000 video samples in the dataset. 
\begin{figure}[t]
    \centering
    \begin{subfigure}[b]{\textwidth}
        \centering
        \includegraphics[width=\textwidth]{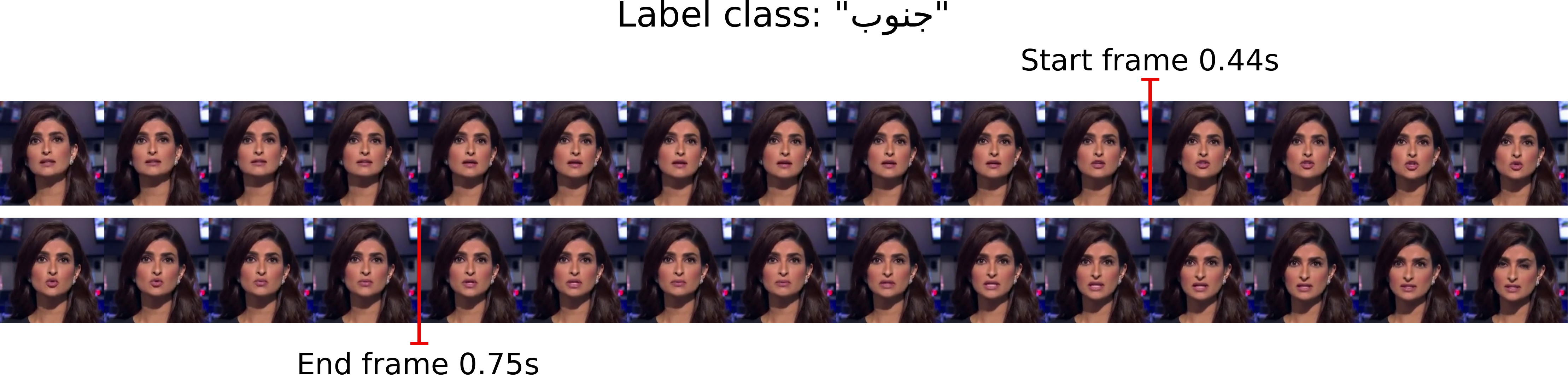}
        \label{fig:sample1}
    \end{subfigure}
    \hfill
    \begin{subfigure}[b]{\textwidth}
        \centering
        \includegraphics[width=\textwidth]{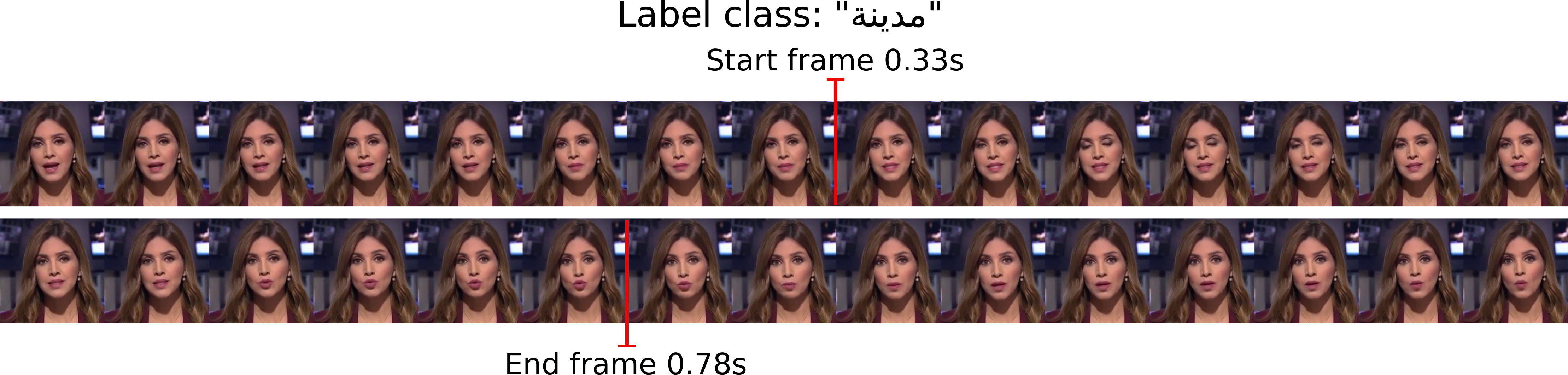}
        \label{fig:sample2}
    \end{subfigure}
    \hfill
    \begin{subfigure}[b]{\textwidth}
        \centering
        \includegraphics[width=\textwidth]{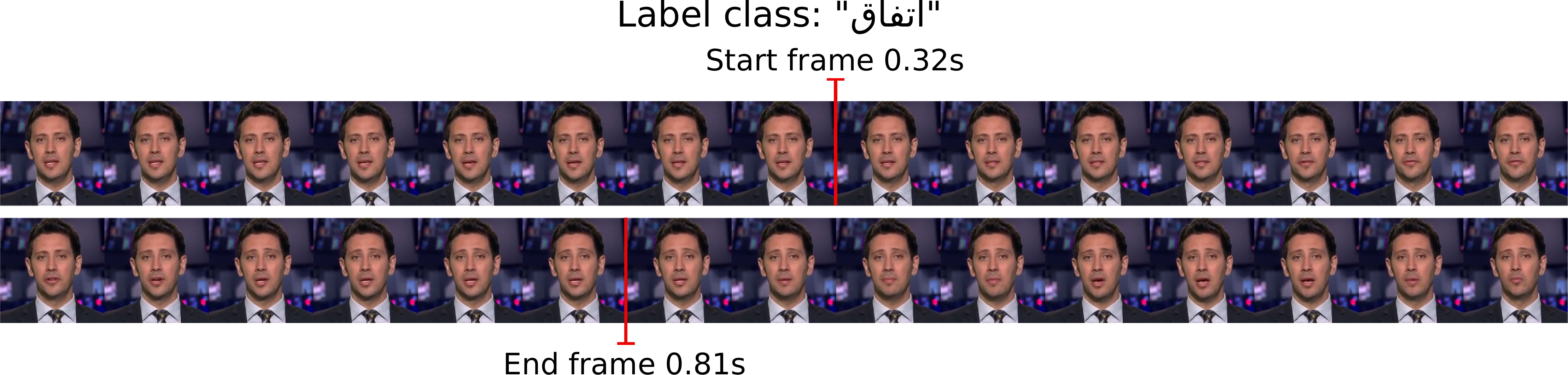}
        \label{fig:sample3}
    \end{subfigure}
    \caption{Three different samples from the LRW-AR dataset. The starting and ending frames for each uttered word are precisely indicated in every sample in the dataset.}
    \label{fig:combined_samples}
\end{figure}

Furthermore, each video sequence in the dataset corresponds to a 1.2 second audio clip captured at a frame rate of 25 frames per second with a video resolution of 256x256. In these sequences, the ground truth word is spoken in the middle, often surrounded by other words or sub-words. This characteristic adds complexity to the lipreading task by simulating more natural speaking conditions.
Each video also comes with an annotation file that specifies the start and end times of the spoken words, ensuring precise alignment for training and evaluation purposes. 
The dataset was partitioned into training, validation, and testing sets. Specifically, 80\% of the samples were designated for training, 10\% for testing, and another 10\% for validation.
To provide a clearer understanding, Figure \ref{fig:combined_samples} showcases samples from the LRW-AR dataset. This figure displays three videos from three different classes, spoken by different speakers, with clear indications of where each corresponding word starts and ends.

%----------------------------------------------------------------------------
\section{Experiments and results}\label{sec5}

In this section, we present various experiments conducted to assess the performance of the proposed lipreading system and its parameters. First, we provide details about our implementation and the datasets used in our experiments. Then, We conduct various experiments as part of our evaluation to demonstrate the precision and efficacy of our system. Specifically, we emphasize the validation of our system in the Arabic language. In this context, we examine the effects of different feature configurations, varying numbers of facial landmarks, and cross-attention heads on the system's accuracy.
Finally, we present a comprehensive analysis outlining the strengths and weaknesses of our lipreading system, along with identified areas for potential improvement.

\subsection{Implementation details} 
Our implementation is based on the PyTorch library \cite{paszke2019pytorch}. In all forthcoming experiments discussed in the remaining section, we trained all architecture configurations using an NVIDIA Titan V GPU with 12GB of RAM. The training process was carried out with a mini-batch size of 32 for a total of 2000 epochs. 
The AdamW optimizer \cite{loshchilov2017decoupled} is employed with an initial learning rate of 3e-4.  We use a cosine annealing approach to reduce the learning rate without requiring a warm-up phase. Additionally, for all trials, we incorporate variable-length augmentation \cite{martinez_lipreading_2020}. The code source is available through the following link: \href{https://github.com/crns-smartvision/lrwar}{https://github.com/crns-smartvision/lrwar}.

\subsection{Datasets}
We conducted experiments using two datasets: our LRW-AR dataset and the ArabicVisual dataset \cite{alsulami2022deep}, a commonly used benchmark dataset for audio-visual speech recognition. It is worth mentioning that ArabicVisual is currently the only lipreading dataset available for the Arabic language, whereas the AVAS and RML datasets are not publicly accessible.

To assess the performance of our lipreading model on the ArabicVisual dataset, we employed a random split of 75\% for training, 15\% for validation, and 10\% for testing, as outlined in \cite{alsulami2022deep}. Accuracy was utilized as the evaluation metric, representing the percentage of correctly classified videos.

\subsection{Arabic lipreading performance evaluation}
To evaluate the effectiveness of the proposed method, we conducted experiments using the LRW-AR dataset, comparing our method with state-of-the-art techniques. These experiments aim to demonstrate our model's performance in word-level lipreading tasks relative to these methods. We have selected the most relevant methods for which the source codes are publicly available:
\begin{itemize}
\item Petridis \etal \cite{petridis2018end} utilized a 3D convolutional neural network (3D Conv) combined with ResNet-34 as the frontend to extract spatial and temporal features. This was followed by a Bi-directional Gated Recurrent Unit (Bi-GRU) network as the backend to capture the temporal dependencies across frames.
\item Zhao \etal \cite{zhao2020mutual} proposed a method that employs a 3D Conv network along with ResNet-18 as the frontend. Similar to \cite{petridis2018end}, the backend consists of a Bi-GRU network.
\item Ma \etal \cite{ma_towards_2021} introduced a method that uses 3D convolution combined with Shufflenetv2 for efficient feature extraction in the frontend. For the backend, a Depthwise Separable Temporal Convolutional Network (TCN) was used. 
\end{itemize}

\begin{table*}[ht]
\center
\centering
% \begin{tabular}{c|c|c|c}
\scalebox{0.825}{

\begin{tabular}{@{}l|l|l|l@{}}
\toprule
Methods & Frontend & Backend & \!Accuracy \\
\midrule
Petridis \etal \cite{petridis2018end} \!\!& 3D Conv \!\!+ \!\!ResNet-34  & Bi-GRU & 78.60\% \\
Zhao \etal \cite{zhao2020mutual}& 3D Conv \!\!+ \!\!ResNet-18  & Bi-GRU & 79.85\% \\
Ma \etal \cite{ma_towards_2021}& 3D Conv \!\!+ \!\!Shufflenetv2  &Depthwise \!Separable \!TCN \!\!\! & 83.23\% \\
Ours & 3D Conv \!\!+ \!\!ResNet-18 \!\!+ \!\!GAT \!\!+ \!\!Tx \!\! & \! MS-TCN \!\!+ \!\!FusionNet & \textbf{85.85\%} \\
\bottomrule
\end{tabular}
}
%\vspace{1ex}
\caption{Comparison of our method with state-of-the-art techniques using the LRW-AR dataset.}
\label{table:accuracy}
\end{table*}

Table \ref{table:accuracy} summarizes the obtained results for the evaluated methods. The final row presents the results of our method, which integrates 3D Conv + ResNet-18 with GAT and Transformer (Tx) in the frontend, and employs MS-TCN + FusionNet in the backend. As shown, our method outperforms other state-of-the-art lipreading techniques achieving the highest accuracy of 85.85\% compared to the accuracies of 78.60\%, 79.85\% and 83.23\% reported by \cite{petridis2018end}, \cite{zhao2020mutual} and \cite{ma_towards_2021}, respectively.

%results on ArabicVisual arabic dataset
Moreover,  we compared our lipreading system with Alsulami \etal \cite{alsulami2022deep} work on the ArabicVisual dataset based on the accuracy metric (percentage of correctly classified videos). Table \ref{table:accuracy_sys} shows the obtained results for phrase classes only, digit classes only, and both digit and phrase classes.

\begin{table*}[ht]
\center
\small

\centering
\begin{tabular}{@{}llll@{}}
\toprule
Methods & Digits & Phrases & Digits \& Phrases \\
\midrule
Alsulami \etal \cite{alsulami2022deep} & \textbf{94\%} & 97\% & 93\%  \\
Ours & 93.75\% & \textbf{100\%} & \textbf{97.77\%} \\
\bottomrule
\end{tabular}
\caption{Comparative results of our lipreading system and the system developed by Alsulami \etal \cite{alsulami2022deep} on ArabicVisual dataset}
\label{table:accuracy_sys}
\end{table*}

Based on the obtained results, our approach appears to have achieved better accuracy than the Alsulami \etal \cite{alsulami2022deep} approach.  As outlined in Table \ref{table:accuracy_sys}, our approach achieved 100\% accuracy for the 'phrases' and 97.77\% for the 'digits \& phrases' classes. However, Alsulami \etal reported accuracy rates of 97\% and 93\%, respectively, for the same classes. Nevertheless, our approach demonstrates competitive performance, with the Alsulami \etal approach only marginally surpassing ours by 0.25\% in digit recognition. These results highlight the limitations of the ArabicVisual dataset \cite{alsulami2022deep} due to its small scale in both size and vocabulary, emphasizing the need for a larger dataset that includes more challenging and representative data.

\subsection{Ablation study}
 \begin{figure*}[t]
\centering\includegraphics[width=0.5\textwidth]{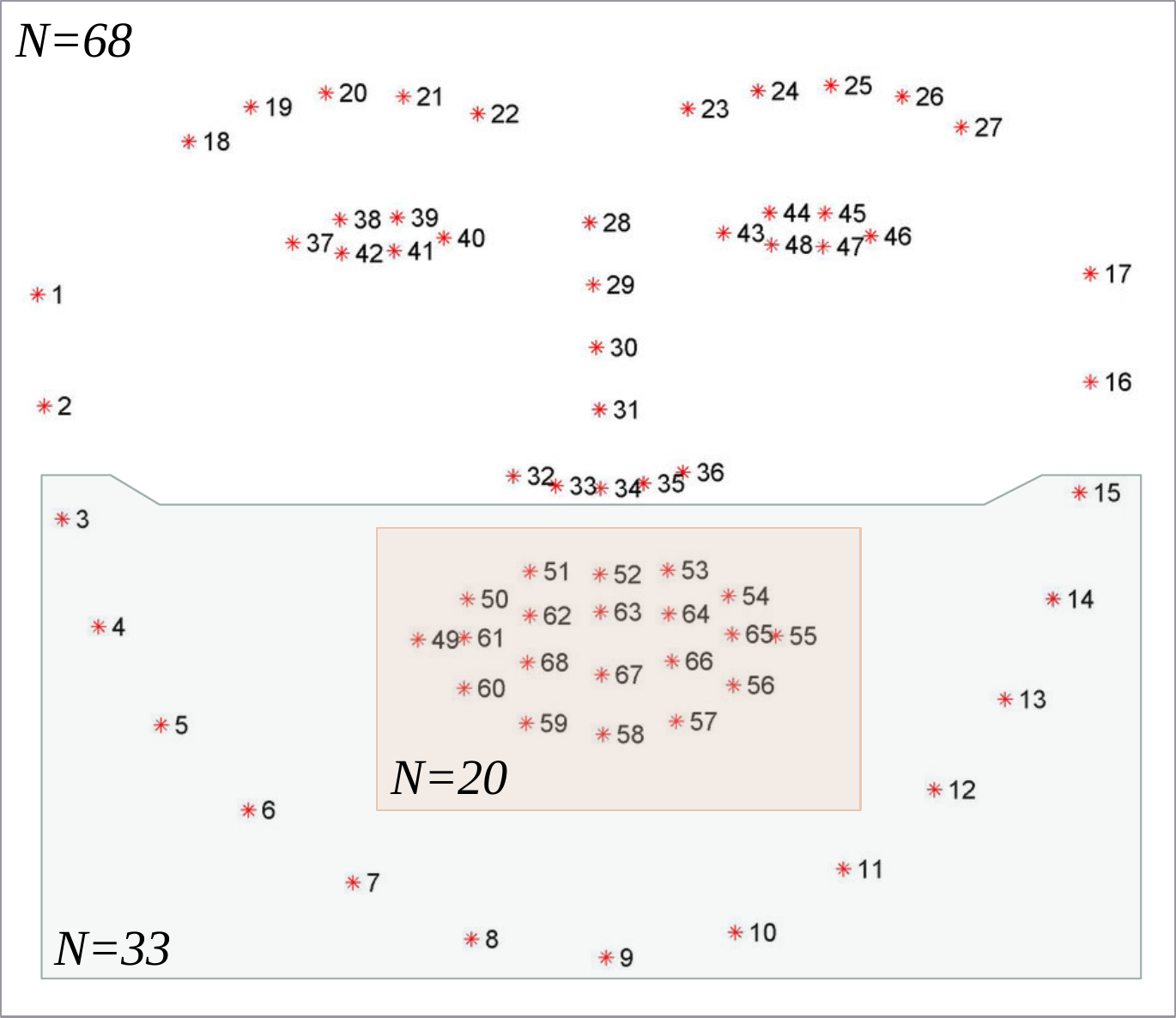}
\caption{The process involves choosing $N$ facial landmark points, which are points used to identify and analyze facial features. The total number of facial landmarks is 68, of which 33 are situated in the lower half of the face (marked by a blue box), and 20 are concentrated in the mouth region (marked by an orange box).}
\label{fig:N}
\end{figure*}

\subsubsection{Facial landmark selection}
In this section, we conducted experiments to assess how the landmark subset selection affects the performance of our system. We evaluated the accuracy of our approach using the standard set of 68 face landmarks, encompassing all facial landmarks, and two subsets of landmarks: 33 points (covering the lower part of the face) and 20 points (covering the lips only). Figure \ref{fig:N} illustrates these three configurations.
The obtained results for these three subsets of landmarks are presented in Table \ref{table:N}.
As indicated in Table \ref{table:N}, employing only 20 landmark points yields the lowest accuracy with 83.8\%. Conversely, utilizing 68 landmark points resulted in improvement, with an accuracy of 84\%.
Furthermore, by employing 33 landmarks, we attained the highest accuracy of 85.85\%.

\begin{table*}[ht]
\center
\small

% \caption{Results on different numbers of landmarks for the full architecture.}
\centering
% \begin{tabular}{c|c|c|c}
\begin{tabular}{@{}llll@{}}
\toprule
&  \multicolumn{3}{c}{Visual and landmarks}\\
\hhline{~---}
& 20 points & 33 points & 68 points \\
\midrule
Accuracy & 83.8\% & \textbf{85.85}\% & 84\%  \\
%84.3 true
%84.5 best test
\bottomrule
\end{tabular}
\caption{Our proposed architecture performance for the three subsets of selected landmarks.}
\label{table:N}
\end{table*}

\begin{figure*}[t]
%\hspace*{-0.25cm} % shift figure 1cm to the left
\centering
\setlength\tabcolsep{6pt} % default: 6pt
\begin{tabular}{ccc}
\includegraphics[width=0.47\linewidth]{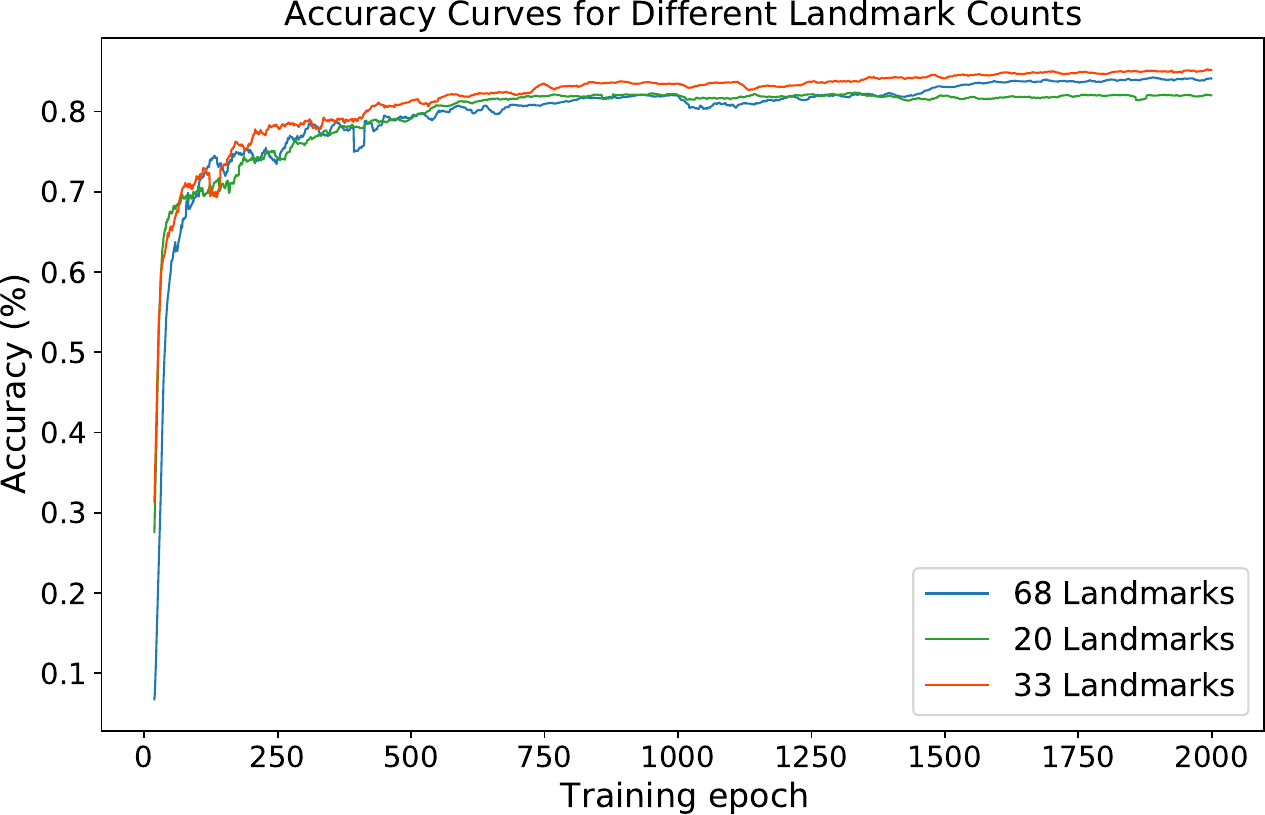} & \includegraphics[width=0.47\linewidth]{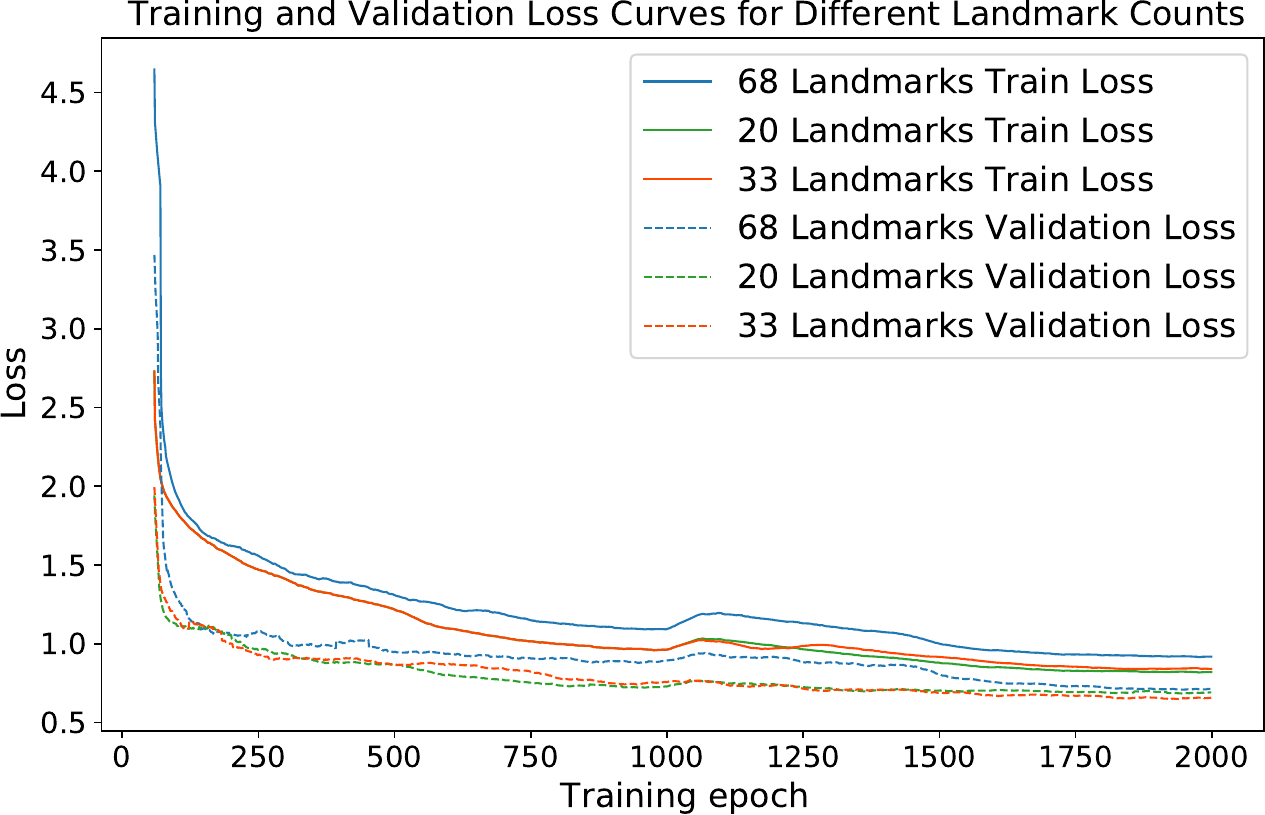}  \\
\begin{subfigure}{0.05\linewidth} \caption{}\label{fig:VO_loss_a} \end{subfigure} & \begin{subfigure}{0.05\linewidth} \caption{} \label{fig:VO_loss_b} \end{subfigure} \\
\end{tabular}
\caption{Accuracy and loss curves. (a) Validation and training accuracy curves, (b) validation and training loss curves.}
\label{fig:loss20}
\end{figure*}

Figure \ref{fig:loss20} illustrates the curves for validation and training accuracy (a) and losses (b) across varying numbers of employed landmarks.  Notably, the optimal accuracy is attained when utilizing 33 landmarks.
Moreover, The loss curves associated with 33 landmarks exhibit superior convergence speed and overall performance.
These results highlight the effectiveness of FusionNet in leveraging landmarks to improve lipreading accuracy, where the 33 landmark configuration represents the optimal choice for our lipreading systems.

\subsubsection{FusionNet number of heads selection}
To enhance the performance of our lipreading system, we evaluated the number of heads ($H$) in the cross-attention fusion mechanism.
Our goal was to investigate the influence of this parameter on our system's ability to recognize spoken words. We conducted evaluations with various values of $H$: 1, 2, 4, 8, and 16.
Table \ref{tab:performance_metrics} summarizes the performance metrics for different values of $H$.

\begin{table}[!h]
\centering

\begin{tabular}{@{}llllll@{}}
\toprule
&  \multicolumn{5}{c}{Number of Heads} \\
\hhline{~-----}
& $H=1$ & $H=2$ & $H=4$ & $H=8$ & $H=16$ \\
\midrule
Accuracy & 83.2\% & 83.85\% & 84.60\% & \textbf{85.85\%} & 82.25\% \\
\bottomrule
\end{tabular}
\caption{Performance metrics for various $H$ values in cross-attention fusion.}
\label{tab:performance_metrics}
\end{table}

Starting with a single head $H=1$, our system achieved an accuracy of 83.2\%, indicating an acceptable baseline performance. As we raised the number of heads to $H=2$, accuracy increased a bit to 83.85\%. This shows that a dual-headed approach slightly enhanced our system's performance. However, the most improvement in performance was achieved when $H=4$, resulting in an accuracy of 84.60\%
This indicates that a moderate level of parallelism and attention capacity is beneficial in decoding spoken words. On the other hand, the performance of our system has been improved, reaching an accuracy of 85,85\% with $H=8$. This configuration has outperformed all other settings in capturing the complex dynamics of cross-modal interactions.
However, increasing the number of heads to $H=16$ lead to a slight drop in accuracy to 82.25\%, demonstrating that an excessive number of heads may add needless complexity.
In summary, our thorough investigation demonstrates the critical role of the number of heads in the cross-attention fusion mechanism. The findings show that, among the evaluated configurations, $H=8$ heads represents the best option, offering the highest level of precision in decoding both visual and geometric features.

\subsubsection{Modality fusion comparison}

In this experiment, our objective was to evaluate the performance of our lipreading system in the Arabic language using five distinct configurations: visual only (VO), landmarks only (LO), concatenated visual and landmarks features (VL[concat]), fused features using single modality attention (VL[SingleAtt]) and our full system configuration using FusionNet (VL[FusionNet]). 

\begin{table}[hb]
\centering
\scalebox{0.825}{
\begin{tabular}{l|c|c|c|c|l}
\toprule
\multirow{2}{*}{Network architectures} &  \multicolumn{4}{c|}{Modality} & \multirow{2}{*}{Accuracy}\\
\hhline{~----~}
 &  {visual} &  {20 landmarks} &  {33 landmarks} &  {68 landmarks}& {} \\
\midrule
{} &  {N/A} & {\checkmark} & {} &  {} &  {59.75\%} \\
LO &  {N/A} &  {} &  {\checkmark} & {} &  {62.33\%} \\
{} &  {N/A} &  {} &  {} & {\checkmark} &  {\underline{65.45\%}} \\
\midrule
VO &  {\checkmark}  & {N/A} &  {N/A} & {N/A} &  {\underline{83.45\%}} \\
\midrule
{} &  {\checkmark} &  {\checkmark} &  {} & {} &  {82.05\%} \\
VL[concat] &  {\checkmark} &  {} &  {\checkmark} & {} &  {83.75\%} \\
{} &  {\checkmark} & {}  &  {} & {\checkmark} &  {\underline{84.25\%}} \\
\midrule
{}&  {\checkmark} &  {\checkmark} &  {} & {}&  {84.20\%} \\
VL[SingleAtt] &  {\checkmark} &  {} & {\checkmark} & {} &   {\underline{85.4}\%}\\
{} &  {\checkmark} & {}  &  {} & {\checkmark} & {83.75\%}\\

\midrule
%\multirow{1}{*}{VL[FusionNet]} 
 & {\checkmark} &  {\checkmark} &  {} & {} &  {83.8\%} \\
VL[FusionNet] &  {\checkmark} &  {} &  {\checkmark} & {} &  {\underline{\textbf{85.85\%}}} \\
{} &  {\checkmark} & {}  &  {} & {\checkmark} & {84.0\%}\\

\bottomrule
\end{tabular}
}
\caption{Performance comparison of different network configurations and modalities on the LRW-AR dataset. The table presents the accuracy achieved with various combinations of visual data and the three subsets of landmarks (\ie 20, 33, and 68)}
\label{tab:configurations}
\end{table}

For the LO configuration, extracted landmarks were forwarded to the geometric-feature network. In the VO configuration, preprocessed data was input into the visual-feature network.
In the combined configuration, we utilized three distinct architectures: the concatenated architecture VL[concat], the VL[SingleAtt] architecture where we used a separate attention mechanism for each modality, and the FusionNet architecture VL[FusionNet], leveraging the synergy between visual and geometric features.

Table \ref{tab:configurations} presents a comparison of various network architectures and their performance on the LRW-AR dataset, demonstrating the effectiveness of different combinations of visual data and landmark configurations. The LO configuration with 68 landmarks achieved an accuracy of 65.45\%, showing the value of detailed geometric features. The VO configuration performed significantly better with an accuracy of 83.45\%, highlighting the importance of visual features. When combining visual data with landmarks, the VL[concat] approach saw the highest accuracy of 84.25\% with 68 landmarks. The VL[SingleAtt] method, integrating visual data with 33 landmarks using a single attention mechanism, achieved a notable accuracy of 85.4\%. However, the VL[FusionNet] method outperformed all other configurations, reaching the highest accuracy of 85.85\% with 33 landmarks, showcasing the superior capability of FusionNet in effectively fusing visual and geometric features for enhanced lipreading accuracy.

\begin{figure*}[!ht]
    \centering
    \begin{subfigure}{0.45\linewidth}
        \includegraphics[width=\linewidth]{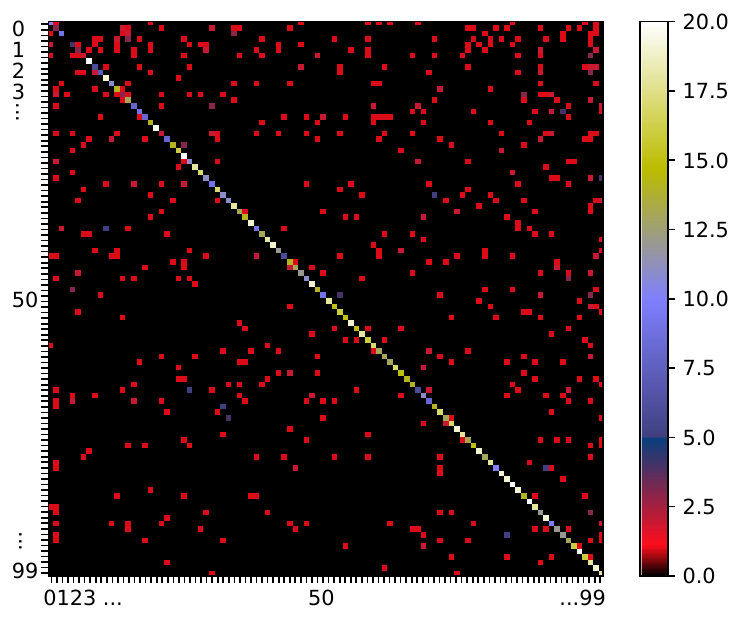}
        \caption{}\label{fig:conf_matrix_a}
    \end{subfigure}
    \hspace{0.01\linewidth} % Add some horizontal space between the subfigures
    \begin{subfigure}{0.45\linewidth}
        \includegraphics[width=\linewidth]{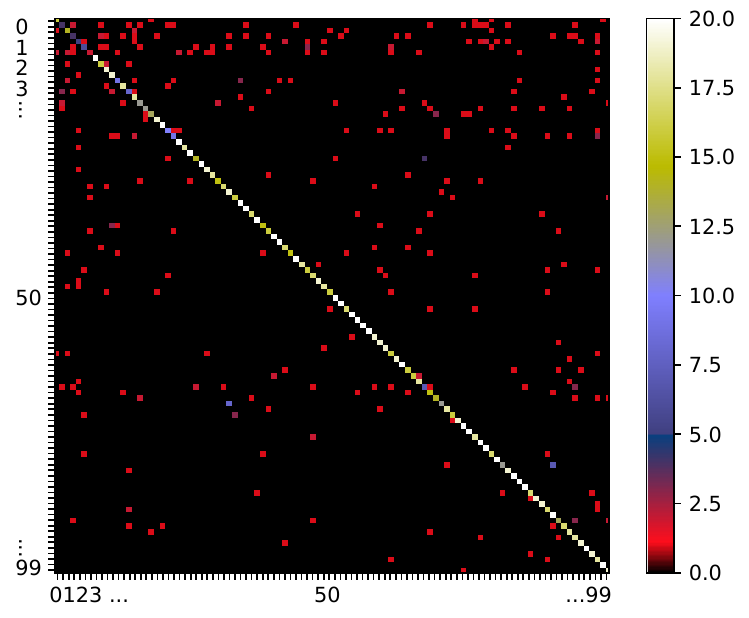}
        \caption{}\label{fig:conf_matrix_b}
    \end{subfigure}
    
    \begin{subfigure}{0.45\linewidth}
        \includegraphics[width=\linewidth]{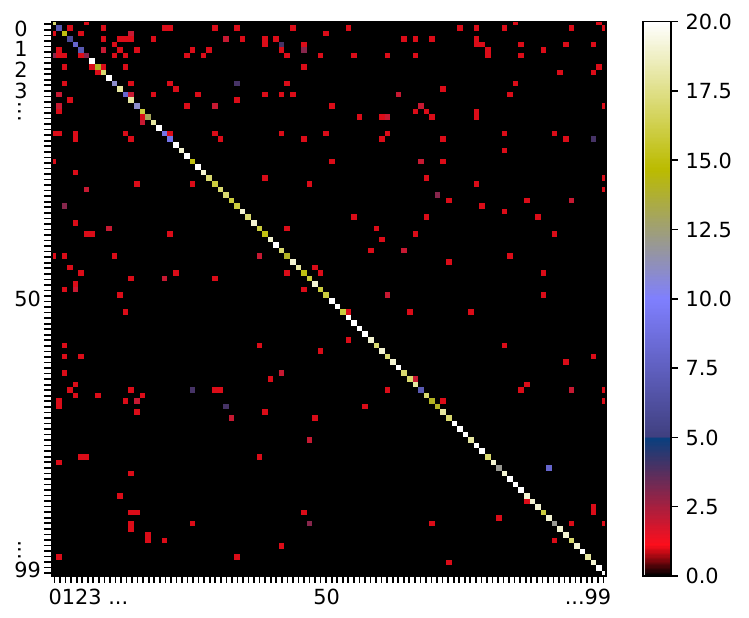}
        \caption{}\label{fig:conf_matrix_c}
    \end{subfigure}
    \hspace{0.01\linewidth} % Add some horizontal space between the subfigures
            \begin{subfigure}{0.45\linewidth}
        \includegraphics[width=\linewidth]{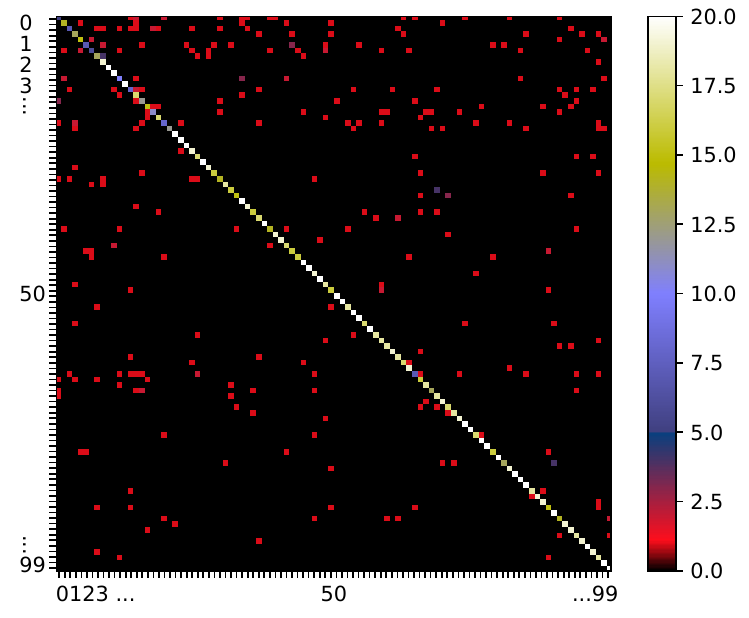}
        \caption{}\label{fig:conf_matrix_d}
    \end{subfigure}
    \begin{subfigure}{0.45\linewidth}
        \includegraphics[width=\linewidth]{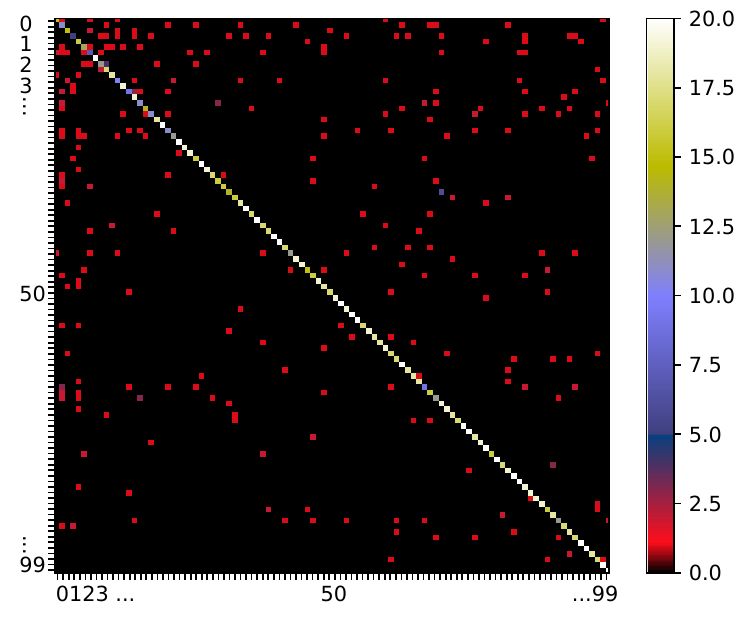}
        \caption{}\label{fig:conf_matrix_e}
    \end{subfigure}

    \caption{Confusion matrices for different models. (a) LO (b) VO (c) VL[concat] (d) VL[SingleAtt] (e) VL[FusionNet].}
\label{fig:conf_matrix}
\end{figure*}

To qualitatively evaluate our system's performance, we present the confusion matrices obtained by different configuration underlined in table \ref{tab:configurations}. In Figure \ref{fig:conf_matrix}, the confusion matrices (\subref{fig:conf_matrix_a}), (\subref{fig:conf_matrix_b}), (\subref{fig:conf_matrix_c}) (\subref{fig:conf_matrix_d}) and (\subref{fig:conf_matrix_e}) present five different configurations: landmarks only (LO), visual only (VO), concatenation architecture VL[concat], VL[SingleAtt] and VL[FusionNet] architecture, respectively. We note that the classes in the confusion matrices are sorted by the number of characters in the word, where all the words are arranged in ascending order based on their length (number of characters), from the shortest to the longest.

The obtained confusion matrices show that the VL[FusionNet] architecture consistently outperforms the other architectures, achieving the highest accuracy across all classes with different word lengths. 
Additionally, VL[concat] and VL[SingleAtt] architecture demonstrated their robustness, especially for shorter word classes, but to a lesser extent.

In contrast, the LO architecture demonstrates reasonable performance, achieving high accuracy for certain longer classes but struggling with shorter ones. 
Nevertheless, the VO architecture, despite surpassing the LO configuration, encounters difficulties with certain shorter classes but performs more effectively with longer ones.

In summary, the VL[FusionNet] architecture excels at recognizing both shorter and longer classes, demonstrating its adaptability and efficiency in handling a large vocabulary. These results highlight the importance of incorporating both visual and landmark information into the system, with FusionNet emerging as the most robust and accurate configuration for Arabic lipreading.

\subsection{Discussion}
Several key observations can be drawn from the experimental results. 
First, we conducted a comprehensive set of experiments to benchmark our approach against state-of-the-art methods in Arabic lipreading systems. Our proposed method demonstrated competitive performance, achieving an accuracy of 85.85\%, which surpasses the results of existing methods that reached 78.60\%, 79.85\%, and 83.23\%, respectively. These results highlight the effectiveness of our approach in improving the accuracy of Arabic lipreading tasks.

For the ablation study, we conducted several experiments to demonstrate the complementarity between geometric features and visual cues, as well as the effectiveness of the proposed cross-attention mechanism in leveraging both modalities for optimal performance. Specifically, we evaluated the impact of different subsets of landmarks (\ie, 20, 33, and 68 landmarks) and the number of heads in the multi-head cross-attention mechanism. Our experiments revealed that employing 8 heads resulted in the highest accuracy, reaching 85.85\%, as previously discussed. These results showed the need for a cross attention mechanism, rather than straightforward feature concatenation and standard attention mechanisms, to achieve the best fusion of geometric and visual features.

In addition to validating our lipreading method, the experiments also aimed to evaluate the constructed LRW-AR dataset. We selected state-of-the-art methods based on the availability of their official implementations, even though these methods were not originally designed for Arabic lipreading. We benchmarked these methods on LRW-AR, and the consistent performance across various models highlights the dataset's robustness and demonstrates that it is not biased toward specific architectures. Furthermore, the uniqueness of LRW-AR lies in being the first large-scale lipreading dataset specifically designed for the Arabic language, marking a significant contribution to the field.

We also intentionally tested another available Arabic lipreading datasets, ArabicVisual \cite{alsulami2022deep}, to highlight their size limitations. Our method achieved 100\% accuracy in phrase recognition on this dataset, indicating that it no longer presents a significant challenge. This underscores the need for larger datasets with more extensive vocabularies. While lipreading datasets for English have received considerable attention, LRW-AR is the first large-scale lipreading dataset specifically for Arabic. It represents a significant advancement in the field and could be continuously expanded in future releases to further bridge the gap with datasets for other languages.

\section{Conclusion}\label{sec6}

In this paper, we introduce an Arabic lipreading system that uses a fusion method of cross-attention mechanisms to decode spoken words from visual data. Our main contribution is to fuse facial landmark data with visual information, improve the accuracy of the system, and effectively address the limitations of lipreading data.
Additionally, we present the larger-scale Arabic LRW-AR dataset to evaluate the system performance.
Experimental results demonstrate the effectiveness of our proposed system in predicting spoken words, achieving the best accuracy result of 85.85\%.
The proposed system represents a significant advancement in the field of visual Arabic speech recognition, highlighting the need for language-specific solutions. In our future work, we aim to extend our system to support sentence-level lipreading. These advancements will enable real-time application development and further improve continuous speech recognition, especially for the Arabic system.

%----------------------------------------------------------------------------
\section*{CRediT author contribution statement}

\textbf{Samar Daou:} Writing – review \& editing, Writing – original draft, Visualization, Methodology, Investigation, Data curation, Software, Conceptualization. \textbf{Achraf Ben-Hamadou:} Writing – review \& editing, Methodology, Conceptualization, Software, Validation, Supervision. \textbf{Ahmed Rekik:} Writing – review \& editing, Formal analysis, Methodology, Conceptualization, Software. \textbf{Abdelaziz Kallel:} Writing – review \& editing, Supervision.

\section*{Declaration of competing interest}

The authors declare that there are no conflicts of interest.

\section*{Data availability}
The dataset generated during the current study is publicly available (\href{https://osf.io/rz49x}{https://osf.io/rz49x}).

\bibliographystyle{elsarticle-num} 
\bibliography{bibliography}

\end{document}